\documentclass[final]{cvpr}

\usepackage{times}
\usepackage{epsfig}
\usepackage{graphicx}
\usepackage{amsmath}
\usepackage{amssymb}
\usepackage{multirow}
\usepackage{sidecap}  
\DeclareMathOperator*{\argmax}{arg\,max}

\usepackage{bm}
\usepackage{multirow}
\usepackage{pifont}
\newcommand{\cmark}{\ding{51}}%
\usepackage{booktabs}
\usepackage{multirow}
\usepackage{subcaption}

\usepackage[pagebackref=true,breaklinks=true,colorlinks,bookmarks=false]{hyperref}

\def\cvprPaperID{955} 
\def\confYear{CVPR 2021}

\begin{document}

\title{Visualizing Adapted Knowledge in Domain Transfer}

\author{Yunzhong Hou \qquad  Liang Zheng\\
	 Australian National University \\
   {\tt\small \{firstname.lastname\}@anu.edu.au}
}


\maketitle


\begin{abstract}

A source model trained on source data and a target model learned through unsupervised domain adaptation (UDA) usually encode different knowledge. To understand the adaptation process, we portray their knowledge difference with image translation. Specifically, we feed a translated image and its original version to the two models respectively, formulating two branches. Through updating the translated image, we force similar outputs from the two branches. When such requirements are met, differences between the two images can compensate for and hence represent the knowledge difference between models. To enforce similar outputs from the two branches and depict the adapted knowledge, we propose a source-free image translation method that generates source-style images using only target images and the two models. We visualize the adapted knowledge on several datasets with different UDA methods and find that generated images successfully capture the style difference between the two domains. For application, we show that generated images enable further tuning of the target model without accessing source data. Code available at \url{https://github.com/hou-yz/DA_visualization}.

\end{abstract}

\section{Introduction}

\begin{figure}
\centering
    \begin{subfigure}[b]{\linewidth}
    \centering
        \includegraphics[width=\textwidth]{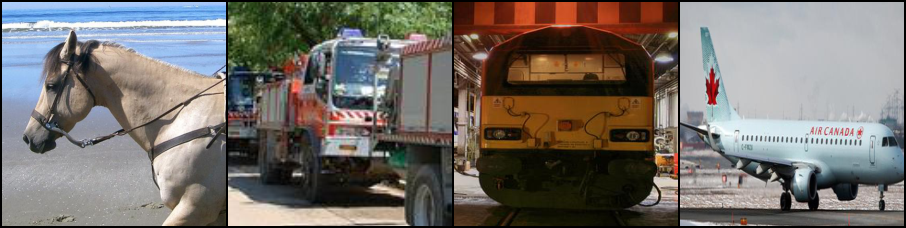}
        \caption{Target images (real-world)}
    \end{subfigure}
    \begin{subfigure}[b]{\linewidth}
    \centering
        \includegraphics[width=\textwidth]{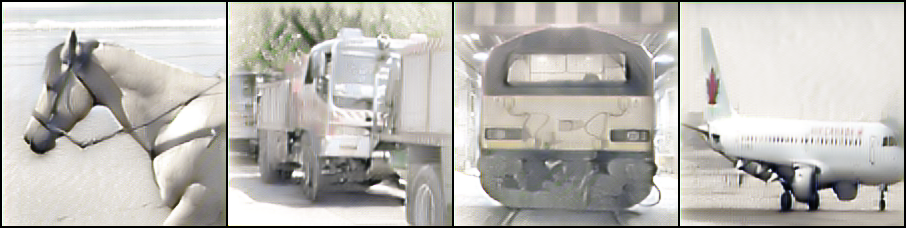}
        \caption{Generated source-style images}
    \end{subfigure}
    \begin{subfigure}[b]{\linewidth}
    \centering
        \includegraphics[width=\textwidth]{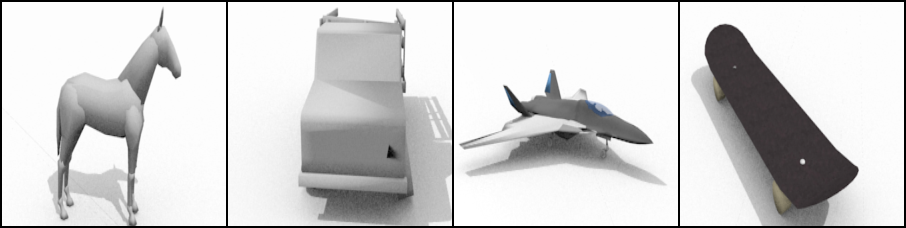}
        \caption{Unseen source images (synthetic)}
    \end{subfigure}
\caption{
Visualization of adapted knowledge in unsupervised domain adaptation (UDA) on the VisDA dataset~\cite{visda2017}. 
To depict the knowledge difference, in our source-free image translation (SFIT) approach, we generate source-style images (b) from target images (a). Instead of accessing source images~(c), the training process is guided entirely by the source and target models, so as to faithfully portray the knowledge difference between them. 
}
\label{fig:intro}
\end{figure}

Domain transfer or domain adaptation aims to bridge the distribution gap between source and target domains. Many existing works study the unsupervised domain adaptation (UDA) problem,  where the target domain is unlabeled~\cite{long2015learning,ganin2016domain,tzeng2017adversarial,bousmalis2017unsupervised,pmlr-v80-hoffman18a}. 
In this process, we are interested in what knowledge neural networks learn and adapt.

Essentially, we should visualize the knowledge difference between models: a source model trained on the source domain, and a target model learned through UDA for the target domain. 
We aim to portray the knowledge difference with image generation. Given a translated image and its original version, we feed the two images to the source and the target model, respectively. It is desired that differences between image pairs can compensate for the knowledge difference between models, leading to similar outputs from the two branches (two images fed to two different models). 
Achieving this, we could also say that the image pair represent the knowledge difference. 

This visualization problem is very challenging and heretofore yet to be studied in the literature. 
It focuses on a relatively understudied field in transfer learning, where we distill knowledge differences from \textit{models} and embed it in generated \textit{images}. 
A related line of works, traditional image translation, generates images in the desired style utilizing content images and style images \cite{gatys2016image,huang2017arbitrary,zhu2017unpaired}, and is applied in pixel-level alignment methods for UDA~\cite{liu2016coupled,bousmalis2016domain,taigman2016unsupervised,pmlr-v80-hoffman18a}. 
However, relying on \textit{images from both domains} to indicate the style difference, such works cannot faithfully portray the knowledge difference between source and target \textit{models}, and are unable to help us understand the adaptation process. 



In this paper, we propose a source-free image translation (SFIT) approach, where we translate target images to the source style without using source images. The exclusion of source images prevents the system from relying on \textit{image} pairs for style difference indication, and ensures that the system only learns from the two \textit{models}. Specifically, we feed translated source-style images to the source model and original target images to the target model, and force similar outputs from these two branches by updating the generator network. 
To this end, we use the traditional knowledge distillation loss and a novel relationship preserving loss, which maintains relative channel-wise relationships between feature maps.
We show that the proposed relationship preserving loss also helps to bridge the domain gap while changing the image style, further explaining the proposed method from a domain adaptation point of view. 
Some results of our method are shown in Fig.~\ref{fig:intro}. We observe that even under the source-free setting, knowledge from the two {models} can still power the style transfer from the target style to the source style (SFIT decreases color saturation and whitens background to mimic the \textit{unseen} source style).

On several benchmarks~\cite{lecun2010mnist,37648,saenko2010adapting,visda2017}, we show that generated images from the proposed SFIT approach 
significantly decrease the performance gap between the two models, suggesting a successful distillation of adapted knowledge. 
Moreover, we find SFIT transfers the image style at varying degrees, when we use different UDA methods on the same dataset. This further verifies that the SFIT visualizations are faithful to the models and that different UDA methods can address varying degrees of style differences. 
For applications, we show that generated images can serve as an additional cue and enable further tuning of target models. This also falls into a demanding setting of UDA, source-free domain adaptation (SFDA)~\cite{kundu2020universal,li2020model,liang2020shot}, where the system has no access to source images. 


\section{Related Work}
\label{sec:related}

\textbf{Domain adaptation} aims to reduce the domain gap between source and target domains. 
Feature-level distribution alignment is a popular strategy  \cite{long2015learning,ganin2016domain,tzeng2017adversarial,saito2018maximum}. Long \etal \cite{long2015learning} use the maximum mean discrepancy (MMD) loss for this purpose. 
Tzeng \etal \cite{tzeng2017adversarial} propose an adversarial method, ADDA, with a loss function based on the generative adversarial network (GAN). 
Pixel-level alignment with image translation is another popular choice in UDA \cite{liu2016coupled,bousmalis2016domain,taigman2016unsupervised,shrivastava2017learning,bousmalis2017unsupervised,pmlr-v80-hoffman18a}. 
Hoffman \etal propose the CyCADA \cite{pmlr-v80-hoffman18a} method based on CycleGAN~\cite{zhu2017unpaired} image translation. 
Other options are also investigated. 
Saito \etal \cite{saito2018maximum} align the task-specific decision boundaries of two classifiers. 
Source-free domain adaptation (SFDA) does \textit{not} use the source data and therefore greatly alleviates the privacy concerns in releasing the source dataset. As an early attempt, 
AdaBN~\cite{li2018adaptive} adapts the statistics of the batch normalization layers in the source CNN to the target domain. 
Li \etal \cite{li2020model} generate images with the same distribution of the target images and use them to fine-tune the classifier. Liang \etal \cite{liang2020shot} fine-tune a label smoothed \cite{muller2019does} source model on the target images. 
To the authors' knowledge, there is still yet to be any visualization that can indicate what models learn during adaptation. 

\textbf{Knowledge distillation} transfers knowledge from a pre-trained teacher model to a student model~\cite{hinton2015distilling}, by maximizing the mutual information between teacher outputs and student outputs. Some existing works consider the relationship between instance or pixels for better distillation performance \cite{tung2019similarity,li2020semantic,park2019relational}.
Instead of distilling teacher knowledge on a given training dataset, data-free knowledge distillation (DFKD)~\cite{lopes2017data,pmlr-v97-nayak19a,chen2019data,micaelli2019zero,haroush2019knowledge,yin2020dreaming} first \emph{generates} training data and then \textit{learns} a student network on this generated dataset. Training data can be generated by aligning feature statistics \cite{lopes2017data,haroush2019knowledge,yin2020dreaming}, enforcing high teacher confidence~\cite{lopes2017data,pmlr-v97-nayak19a,chen2019data,haroush2019knowledge,yin2020dreaming}, and adversarial generation of hard examples for the student~\cite{micaelli2019zero,yin2020dreaming}. 
In \cite{haroush2019knowledge,yin2020dreaming}, batch normalization statistics are matched as regularization. 
Our work, while also assuming no access to source images, differs significantly from these works in that our image translation has to portray the transferred knowledge, whereas data-free knowledge distillation just generates whatever images that satisfy the teacher networks. 

\textbf{Image translation} renders the same content in a different artistic style. Some existing works adopt a GAN-based system for this task~\cite{liu2016coupled,taigman2016unsupervised,isola2017image,zhu2017unpaired,pmlr-v80-hoffman18a}, while others use a pre-trained feature extractor for style transfer~\cite{gatys2016image,johnson2016perceptual,luan2017deep,huang2017arbitrary}. 
Zhu \etal adopt a cycle consistency loss in the image translation loop to train the CycleGAN system \cite{zhu2017unpaired}. Gatys \etal consider a content loss on high-level feature maps, and a style loss on feature map statistics for style transfer \cite{gatys2016image}. 
Huang and Belongie \cite{huang2017arbitrary} propose a real-time AdaIN style transfer method by changing the statistics in instance normalization layers. Based on AdaIN, Karras \etal propose StyleGAN for state-of-the-art image generation \cite{karras2019style}. 
Our work differs from traditional image translations in that rather than \textit{images} from the two domains,  only \textit{models} from two domains are used to guide the image update. 


\begin{figure*}[t]
\centering
\includegraphics[width=0.95\linewidth]{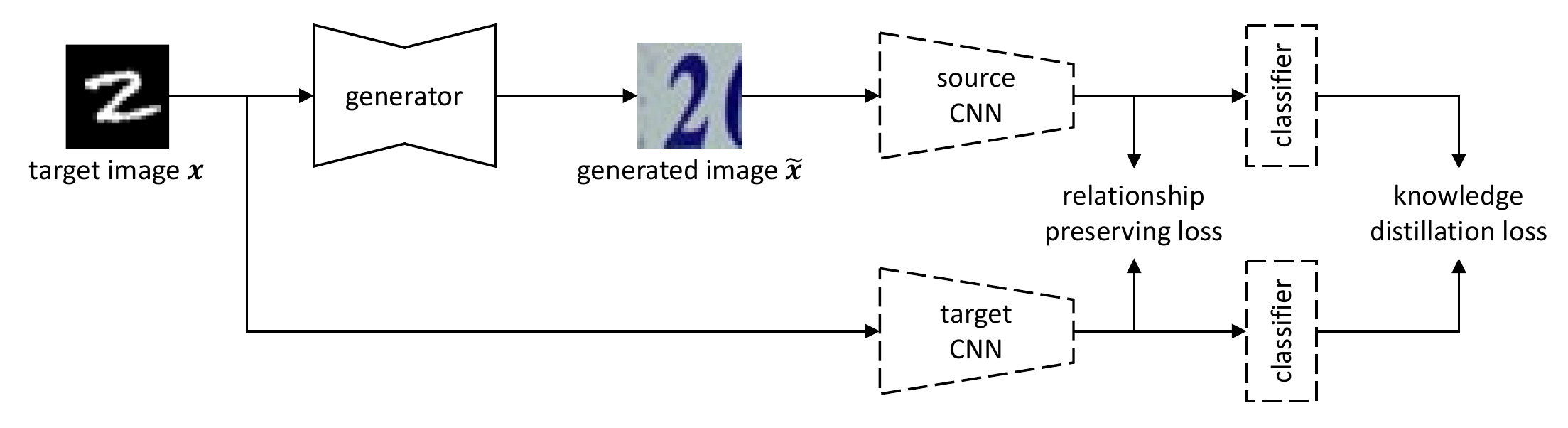}
\caption{
The proposed source-free image translation (SFIT) method for visualizing the adapted knowledge in UDA. 
The system includes two branches: original target images are fed to the target CNN, whereas generated source-style images are fed to the source CNN.
We minimize the knowledge distillation loss and the relationship preserving loss, and update the generator network accordingly. 
If the two branches get similar results while adopting different models, then the difference between the original target image $\bm{x}$ and the generated source-style image $\widetilde{\bm{x}}$ should be able to \textit{mitigate} and therefore \textit{exhibit} the knowledge difference between models. 
Dashed lines indicate fixed network parameters.
}
\label{fig:train_generator}
\end{figure*}

\section{Problem Formulation}
\label{sec:formulation}

To achieve our goal, \emph{i.e.,} visualizing adapted knowledge in UDA, we translate a image $\bm{x}$ from a certain domain to a new image $\widetilde{\bm{x}}$. It is hoped that feeding the original image to its corresponding model (trained for that certain domain) and the generated image to the other model can minimize the output difference between these two branches. The update process is directed only by the source model $f_\text{S}\left(\cdot\right)$ and the target model $f_\text{T}\left(\cdot\right)$, and we prevent access to the images from the other domain to avoid distractions. We formulate the task of visualizing adapted knowledge as a function of the source model, the target model, and the image from a certain domain,
\begin{align}
\label{eq:goal}
    \mathcal{G}\left(f_\text{S},f_\text{T},\bm{x}\right) \rightarrow \widetilde{\bm{x}}.
\end{align}
In contrast, traditional image translation needs access to images from both domains for content and style specification. In addition to the source image $\bm{x}_\text{S}$ and the target image $\bm{x}_\text{T}$, traditional image translation also relies on certain neural network $d\left(\cdot\right)$ as the criterion. Instead of the source and target models, ImageNet~\cite{imagenet_cvpr09} pre-trained VGG \cite{simonyan2014very} and adversarially trained discriminator networks are used for this task in style transfer~\cite{gatys2016image,huang2017arbitrary} and GAN-based methods~\cite{zhu2017unpaired,pmlr-v80-hoffman18a}, respectively. Traditional image translation task can thus be formulated as,
\begin{align}
\label{eq:traditional_image_trans}
    \mathcal{G}\left(d,\bm{x}_\text{S},\bm{x}_\text{T}\right) \rightarrow \widetilde{\bm{x}}.
\end{align}
Comparing our goal in Eq.~\ref{eq:goal} and traditional image translation in Eq.~\ref{eq:traditional_image_trans}, we can see a clear gap between them. Traditional image translation learns the style difference indicated by \textit{images} from both domains, whereas our goal is to learn to visualize the knowledge difference between the source and target \textit{models} $f_\text{S}\left(\cdot\right),f_\text{T}\left(\cdot\right)$.

\section{Method}
\label{sec:method}

To investigate what neural networks learn in domain adaptation, we propose source-free image translation (SFIT), a novel method that generates source-style images from original target images, so as to mitigate and represent the knowledge difference between models. 

\subsection{Overview}
\label{sec:sec:overview}
Following many previous UDA works \cite{ganin2016domain,long2015learning,tzeng2017adversarial,liang2020shot}, we assume that only the feature extractor CNN in the source model is adapted to the target domain. Given a source CNN $f_\text{S}\left(\cdot\right)$ and a target CNN $f_\text{T}\left(\cdot\right)$ sharing the same classifier $p\left(\cdot\right)$, we train a generator $g\left(\cdot\right)$ for the SFIT task. We discuss why we choose this translation direction in Section~\ref{sec:sec:discussion}. 
As the training process is source-free, for simplicity, we refer to the target image as $\bm{x}$ instead of $\bm{x}_\text{T}$ in what follows. 

As shown in Fig.~\ref{fig:train_generator}, given a generated image $\widetilde{\bm{x}} = g\left(\bm{x}\right)$, the source model outputs a feature map $f_\text{S}\left(\widetilde{\bm{x}}\right)$ 
and a probability distribution $p\left(f_\text{S}\left(\widetilde{\bm{x}}\right)\right)$ over all $C$ classes. 
To depict the adapted knowledge in the generated image, in addition to the traditional knowledge distillation loss, 
we introduce a novel relationship preserving loss, which maintains relative channel-wise relationships between the target-image-target-model feature map $f_\text{T}\left(\bm{x}\right)$ and the generated-image-source-model feature map $f_\text{S}\left(\widetilde{\bm{x}}\right)$. 

\subsection{Loss Functions}
\label{sec:sec:loss}


With a knowledge distillation loss $\mathcal{L}_\text{KD}$ and a relationship preserving loss $\mathcal{L}_\text{RP}$, we have the overall loss function,
\begin{equation}
\begin{split}
\label{eq:overall}
    \mathcal{L} = \mathcal{L}_\text{KD} + \mathcal{L}_\text{RP}. 
\end{split}
\end{equation}
In the following sections, we detail the loss terms. 

\textbf{Knowledge distillation loss.}
In the proposed source-free image translation method, portraying the adapted knowledge in the target model $f_\text{T}\left(\cdot\right)$ with source model and generator combined $f_\text{S}\left(g\left(\cdot\right)\right)$ can be regarded as a special case of knowledge distillation, where we aim to distill the adapted knowledge to the generator. 
In this case, we include a knowledge distillation loss between generated-image-source-model output $p\left(f_\text{S}\left(\widetilde{\bm{x}}\right)\right)$ and target-image-target-model output $p\left(f_\text{T}\left(\bm{x}\right)\right)$,
\begin{align}
\label{eq:kd}
    \mathcal{L}_\text{KD} = \mathcal{D}_\text{KL}\left(p\left(f_\text{T}\left(\bm{x}\right)\right), p\left(f_\text{S}\left(\widetilde{\bm{x}}\right)\right)\right),
\end{align}
where $\mathcal{D}_\text{KL}\left(\cdot, \cdot\right)$ denotes the Kullback-Leibler divergence.

\textbf{Relationship preserving loss.}
Similar classification outputs indicate a successful depiction of the target model knowledge on the generated images. 
As we assume a fixed classifier for UDA, the global feature vectors from the target image target CNN and the generated image source CNN should be similar after a successful knowledge distillation. 
Promoting similar channel-wise relationships between feature maps $f_\text{T}\left(\bm{x}\right)$ and $f_\text{S}\left(\widetilde{\bm{x}}\right)$ helps to achieve this goal. 

Previous knowledge distillation works preserve relative batch-wise or pixel-wise relationships \cite{tung2019similarity,li2020semantic}. However, they are not suitable here for the following reasons. Relative batch-wise relationships can not effectively supervise the per-image generation task. Besides, the efficacy of pixel-wise relationship preservation can be overshadowed by the global pooling before the classifier. By contrast, channel-wise relationships are computed on a per-image basis, and are effective even after global pooling. 
As such, we choose the channel-wise relationship preserving loss that is computed in the following manner. 

Given feature maps $f_\text{T}\left(\bm{x}\right),f_\text{S}\left(\widetilde{\bm{x}}\right)$, we first reshape them into feature vectors $\mathcal{F}_\text{S}$ and $\mathcal{F}_\text{T}$,
\begin{equation}
\begin{aligned}
\label{eq:reshape}
     f_\text{S}\left(\widetilde{\bm{x}}\right) \in \mathbb{R}^{D \times H \times W} \rightarrow \mathcal{F}_\text{S}\in \mathbb{R}^{D \times H W},\\
     f_\text{T}\left(\bm{x}\right) \in \mathbb{R}^{D \times H \times W} \rightarrow \mathcal{F}_\text{T}\in \mathbb{R}^{D \times H W},
\end{aligned}
\end{equation}
where $D, H$, and $W$ are the feature map depth (number of channels), height, and width, respectively. Next, we calculate their channel-wise self correlations, or Gram matrices, 
\begin{align}
\label{eq:gram}
     G_\text{S} = \mathcal{F}_\text{S}\cdot\mathcal{F}_\text{S}^T, \;
     G_\text{T} = \mathcal{F}_\text{T}\cdot\mathcal{F}_\text{T}^T,
\end{align}
where $G_\text{S}, G_\text{T} \in \mathbb{R}^{D \times D}$. Like other similarity preserving losses for knowledge distillation \cite{tung2019similarity,li2020semantic}, we then apply the row-wise $\mathcal{L}_2$ normalization,
\begin{align}
\label{eq:l2norm}
     \widetilde{G}_{\text{S}\left[i,:\right]} = \frac{G_{\text{S}\left[i,:\right]}}{\left\|G_{\text{S}\left[i,:\right]} \right\|_2}, \;
     \widetilde{G}_{\text{T}\left[i,:\right]} = \frac{G_{\text{T}\left[i,:\right]}}{\left\|G_{\text{T}\left[i,:\right]} \right\|_2},
\end{align}
where $\left[i,:\right]$ indicates the $i$-th row in a matrix. 
At last, we define the relationship preserving loss as mean square error (MSE) between the normalized Gram matrices,
\begin{align}
\label{eq:relationship}
     \mathcal{L}_\text{RP} = \frac{1}{D}  \left\|\widetilde{G}_\text{S} - \widetilde{G}_\text{T} \right\|_F^2,
\end{align}
where $\left\|\cdot\right\|_F$ denotes the Frobenius norm (entry-wise $\mathcal{L}_2$ norm for matrix). 
In Section~\ref{sec:sec:discussion}, we further discuss the relationship preserving loss from the viewpoint of style transfer and domain adaptation, and show it can align feature map distributions in a similar way as style loss \cite{gatys2016image} for style transfer and MMD loss \cite{long2015learning} for UDA, forcing the generator to portray the knowledge difference between the two models.  

\begin{figure}
\centering
    \begin{subfigure}[b]{0.48\linewidth}
    \centering
        \includegraphics[width=\textwidth]{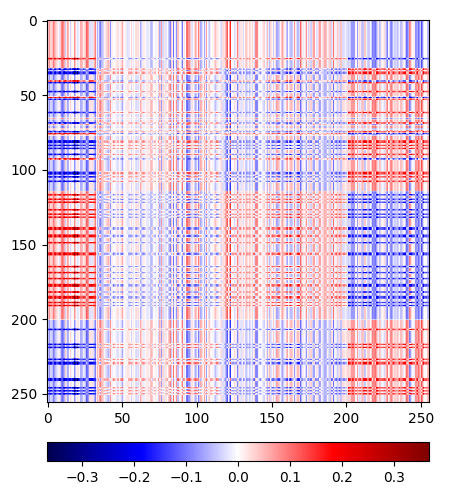}
        \caption{Relationship preserving loss}
    \end{subfigure}
    \hfill
    \begin{subfigure}[b]{0.48\linewidth}
    \centering
        \includegraphics[width=\textwidth]{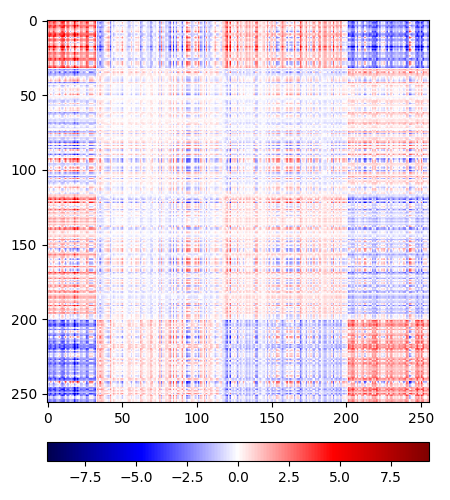}
        \caption{Traditional style loss}
    \end{subfigure}
\caption{Comparison between the proposed relationship preserving loss and the traditional style loss. In (a) and (b), given 256-dimensional feature maps, we show differences of row-wise normalized Gram matrix (Eq.~\ref{eq:relationship}) and original Gram matrix (Eq.~\ref{eq:TS}). Deeper colors indicate larger differences and therefore stronger supervision. The proposed relationship preserving loss provides evenly distributed supervision for all channels, whereas the traditional style loss focuses primarily on several channels. 
}
\label{fig:loss}
\end{figure}

\subsection{Discussions}
\label{sec:sec:discussion}

\textbf{Why transfer target images to the source style.} According to the problem formulation in Eq.~\ref{eq:goal}, we should be able to visualize the adapted knowledge by generating either source-style images from target images, or target-style images from source images. In this paper, we select the former direction as it might be further applied in fine-tuning the target model (see Section~\ref{sec:sec:applications} for application). 

\textbf{Style transfer with the relationship preserving loss.}
The proposed relationship preserving loss can be regarded as a normalized version of the traditional style loss introduced by Gatys \etal \cite{gatys2016image},
\begin{align}
\label{eq:TS}
     \mathcal{L}_\text{style} = \frac{1}{D^2} \left\|{G}_\text{S} - {G}_\text{T} \right\|_F^2,
\end{align}
which computes MSE between Gram matrices. 

In the proposed relationship preserving loss (Eq.~\ref{eq:relationship}), instead of original Gram matrices, we use a row-wise normalized version. It focuses on \textit{relative} relationships between channels, rather than \textit{absolute} values of self correlations as in the traditional style loss.
Preserving relative relationships provides more evenly-distributed supervision for all channels, instead of prioritizing several channels as in the traditional style loss (Fig.~\ref{fig:loss}). 
Experiments find this evenly-distributed supervision better preserves the foreground object and allows for easier training and higher performance, while also changing the image style (see Section~\ref{sec:sec:variants}).

\textbf{Distribution alignment with the relationship preserving loss.}
As proved by Li \etal \cite{10.5555/3172077.3172198}, the traditional style loss $\mathcal{L}_\text{style}$ is equivalent to the MMD loss~\cite{long2015learning} for UDA. 
We can also see the relationship preserving loss as a modified version of the MMD loss, which aligns the distribution of the generated image source CNN feature map $f_\text{S}\left(\widetilde{\bm{x}}\right)$ to the target image target CNN feature map $f_\text{T}\left({\bm{x}}\right)$. 




\section{Experiments}
\label{sec:experiment}

\subsection{Datasets}
We visualize the knowledge difference between source and target models on the following datasets. 

\textbf{Digits} is a standard UDA benchmark that focuses on 10-class digit recognition. Specifically, we experiment on MNIST~\cite{lecun2010mnist}, USPS, and SVHN~\cite{37648} datasets. 

\textbf{Office-31} \cite{saenko2010adapting} is a standard benchmark for UDA that contains 31 classes from three distinct domains: Amazon (A), Webcam (W), and DSLR (D). 

\textbf{VisDA} \cite{visda2017} is a challenging large-scale UDA benchmark for domain adaptation from 12 classes of synthetic CAD model images to real-world images in COCO \cite{lin2014microsoft}. 



\subsection{Implementation Details}
\label{sec:sec:implementations}

\textbf{Source and target models.}
We adopt source and target models from a recent SFDA work SHOT-IM~\cite{liang2020shot} if not specified. SFDA is a special case of UDA, and it is even more interesting to see what machines learn in the absence of source data. We also include UDA methods DAN~\cite{long2015learning} and ADDA~\cite{tzeng2017adversarial} for SFIT result comparisons. 
For network architectures, on digits dataset, following Long \etal~\cite{long2018conditional}, we choose a LeNet \cite{lecun1998gradient} classifier. 
On Office-31 and VisDA, we choose ResNet-50 and ResNet-101~\cite{he2016deep}, respectively. 



\textbf{Generator for SFIT.}
We use a modified CycleGAN \cite{zhu2017unpaired} architecture with 3 residue blocks due to memory concerns. 

\textbf{Training schemes.}
During training, we first initialize the generator as a transparent filter, which generates images same as the original input. To this end, we use the ID loss $\mathcal{L}_\text{ID}=\left\| \widetilde{\bm{x}} - \bm{x} \right\|_1$ and the content loss $\mathcal{L}_\text{content} = \left\| f_\text{S}\left(\widetilde{\bm{x}}\right) - f_\text{S}\left(\bm{x}\right) \right\|_2$  to train the generator for initialization. The initialization performance is shown in Table~\ref{tab:generation_variants}, where we can see a mild 1.9\% accuracy drop from original target images. 
Then, we train the generator with the overall loss function in Eq.~\ref{eq:overall} for visualizing the adapted knowledge. Specifically, we use an Adam optimizer with a cosine decaying \cite{loshchilov2016sgdr} learning rate starting from $3\times10^{-4}$ and a batch size of $16$. 
All experiments are finished using one RTX-2080Ti GPU.

\subsection{Evaluation}
\label{sec:sec:evaluation}

\begin{table}[t]
\resizebox{\linewidth}{!}{
\centering
\begin{tabular}{l|ccc}
\toprule
Method           & SVHN$\rightarrow$MNIST & USPS$\rightarrow$MNIST & MNIST$\rightarrow$USPS \\ \hline
Source only \cite{pmlr-v80-hoffman18a}     & 67.1$\pm$0.6   & 69.6$\pm$3.8   & 82.2$\pm$0.8   \\ 
DAN \cite{long2015learning}             & 71.1       & -          & 81.1       \\ 
DANN \cite{ganin2016domain}            & 73.8       & 73         & 85.1       \\ 
CDAN+E \cite{long2018conditional}             & 89.2   & 98.0   & 95.6   \\  
CyCADA  \cite{pmlr-v80-hoffman18a}          & 90.4$\pm$0.4   & 96.5$\pm$0.1   & 95.6$\pm$0.4  \\ 
MCD \cite{saito2018maximum}             & 96.2$\pm$0.4   & 94.1$\pm$0.3   & 94.2$\pm$0.7   \\ 
GTA \cite{sankaranarayanan2018generate}             & 92.4$\pm$0.9   & 90.8$\pm$1.3   & 95.3$\pm$0.7   \\ 
3C-GAN \cite{li2020model}             & 99.4$\pm$0.1  & 99.3$\pm$0.1   & 97.3$\pm$0.2   \\  \hline
Source model \cite{liang2020shot}    & 72.3$\pm$0.5   & 90.5$\pm$1.6   & 72.7$\pm$2.3   \\ 
Target model \cite{liang2020shot}     & 98.8$\pm$0.1   & 98.1$\pm$0.5   & 97.9$\pm$0.2   \\ 
Generated images & 98.6$\pm$0.1   & 97.4$\pm$0.3   & 97.6$\pm$0.3   \\ \bottomrule
\end{tabular}
}
\caption{Classification accuracy (\%) on digits datasets. 
In Table~\ref{tab:digits}-\ref{tab:visda}, ``Generated images'' refers to  feeding images generated by SFIT to the source model. }
\label{tab:digits}
\end{table}

\begin{table}[t]
\resizebox{\linewidth}{!}{
\begin{tabular}{l|cccccc|c}
\toprule
Method           & A$\rightarrow$W  & D$\rightarrow$W  & W$\rightarrow$D  & A$\rightarrow$D  & D$\rightarrow$A  & W$\rightarrow$A  & Avg. \\\hline
ResNet-50 \cite{he2016deep}       & 68.4 & 96.7 & 99.3 & 68.9 & 62.5 & 60.7 & 76.1 \\
DAN \cite{long2015learning}             & 80.5 & 97.1 & 99.6 & 78.6 & 63.6 & 62.8 & 80.4 \\
DANN \cite{ganin2016domain}            & 82.6 & 96.9 & 99.3 & 81.5 & 68.4 & 67.5 & 82.7 \\
ADDA \cite{tzeng2017adversarial}            & 86.2 & 96.2 & 98.4 & 77.8 & 69.5 & 68.9 & 82.9 \\
JAN \cite{long2017deep}             & 86.0 & 96.7 & 99.7 & 85.1 & 69.2 & 70.7 & 84.6 \\
CDAN+E \cite{long2018conditional}          & 94.1 & 98.6 & 100.0  & 92.9 & 71.0 & 69.3 & 87.7 \\
GTA \cite{sankaranarayanan2018generate}             & 89.5 & 97.9 & 99.8 & 87.7 & 72.8 & 71.4 & 86.5 \\
3C-GAN \cite{li2020model}          & 93.7 & 98.5 & 99.8 & 92.7 & 75.3 & 77.8 & 89.6 \\\hline
Source model \cite{liang2020shot}    & 76.9 & 95.6 & 98.5 & 80.3 & 60.6 & 63.4 & 79.2 \\
Target model \cite{liang2020shot}    & 90.8 & 98.4 & 99.9 & 88.8 & 73.6 & 71.7 & 87.2 \\
Generated images & 89.1 & 98.1 & 99.9 & 87.3 & 69.8 & 68.7 & 85.5 \\
Fine-tuning & 91.8 & 98.7 & 99.9 & 89.9 & 73.9 & 72.0 & 87.7 \\
\bottomrule
\end{tabular}
}
\caption{Classification accuracy (\%) on the Office-31 dataset. In Table~\ref{tab:office} and Table~\ref{tab:visda}, ``Fine-tuning'' refers to target model fine-tuning result with both generated images and target images (see Section \ref{sec:sec:applications} for more details).}
\label{tab:office}
\end{table}

\begin{table}
\centering
\resizebox{\linewidth}{!}{
\setlength{\tabcolsep}{0.5mm}{
\begin{tabular}{l|cccccccccccc|c}
\toprule
Method           & plane & bcycl & bus  & car  & horse & knife & mcycl & person & plant & sktbrd & train & truck & per-class \\\hline
ResNet-101 \cite{he2016deep}       & 55.1  & 53.3  & 61.9 & 59.1 & 80.6  & 17.9  & 79.7  & 31.2   & 81.0    & 26.5   & 73.5  & 8.5   & 52.4      \\
DAN \cite{long2015learning}             & 87.1  & 63.0    & 76.5 & 42.0   & 90.3  & 42.9  & 85.9  & 53.1   & 49.7  & 36.3   & 85.8  & 20.7  & 61.1      \\
DANN \cite{ganin2016domain}            & 81.9  & 77.7  & 82.8 & 44.3 & 81.2  & 29.5  & 65.1  & 28.6   & 51.9  & 54.6   & 82.8  & 7.8   & 57.4      \\
JAN \cite{long2017deep}              & 75.7  & 18.7  & 82.3 & 86.3 & 70.2  & 56.9  & 80.5  & 53.8   & 92.5  & 32.2   & 84.5  & 54.5  & 65.7      \\
ADDA \cite{tzeng2017adversarial}              & 88.8  & 65.7  & 85.6 & 53.1 & 74.9  & 96.2  & 83.3  & 70.7   & 75.9  & 26.4   & 83.9  & 32.4  & 69.7      \\
MCD \cite{saito2018maximum}             & 87.0    & 60.9  & 83.7 & 64.0   & 88.9  & 79.6  & 84.7  & 76.9   & 88.6  & 40.3   & 83.0    & 25.8  & 71.9      \\
CDAN+E \cite{long2018conditional}           & 85.2  & 66.9  & 83.0   & 50.8 & 84.2  & 74.9  & 88.1  & 74.5   & 83.4  & 76.0     & 81.9  & 38.0    & 73.9      \\
SE \cite{french2018selfensembling}              & 95.9  & 87.4  & 85.2 & 58.6 & 96.2  & 95.7  & 90.6  & 80.0     & 94.8  & 90.8   & 88.4  & 47.9  & 84.3      \\
3C-GAN \cite{li2020model}           & 94.8  & 73.4  & 68.8 & 74.8 & 93.1  & 95.4  & 88.6  & 84.7   & 89.1  & 84.7   & 83.5  & 48.1  & 81.6      \\ \hline
Source model \cite{liang2020shot}    & 58.3  & 17.6  & 54.2 & 69.9 & 64.4  & 5.5   & 82.2  & 30.7   & 62.2  & 24.6   & 86.2  & 6.0   & 46.8      \\
Target model \cite{liang2020shot}     & 92.5  & 84.7  & 81.3 & 54.6 & 90.5  & 94.7  & 80.9  & 79.1   & 90.8  & 81.5   & 87.9  & 50.1  & 80.7      \\
Generated images & 88.9  & 65.8  & 83.0 & 61.7 & 88.5  & 76.8  & 89.5  & 69.6   & 91.4  & 51.9   & 84.3  & 34.3  & 73.8      \\
Fine-tuning      & 94.3  & 79.0  & 84.9 & 63.6 & 92.6  & 92.0  & 88.4  & 79.1   & 92.2  & 79.8   & 87.6  & 43.0  & 81.4     \\ \bottomrule   
\end{tabular}
}
}
\caption{Classification accuracy (\%) on the VisDA dataset. }
\label{tab:visda}
\end{table}

\begin{figure}
\centering
    \begin{subfigure}[b]{\linewidth}
    \centering
        \includegraphics[width=\textwidth]{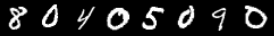}
        \caption{Target images (MNIST)}
    \end{subfigure}
    
    \begin{subfigure}[b]{\linewidth}
    \centering
        \includegraphics[width=\textwidth]{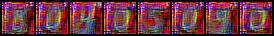}
        \caption{Generated source-style images}
    \end{subfigure}
    
    \begin{subfigure}[b]{\linewidth}
    \centering
        \includegraphics[width=\textwidth]{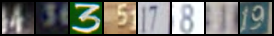}
        \caption{Unseen source images (SVHN)}
    \end{subfigure}
\caption{
Results from the SFIT method on digits datasets SVHN$\rightarrow$MNIST. 
In Fig.~\ref{fig:intro} and Fig.~\ref{fig:generation_results_digits}-\ref{fig:generation_results_visda}, we show in (a): target images, (b): generated source-style images, each of which corresponds to the target image above it, and (c): the unseen source images. 
For gray-scale target images from MNIST, our SFIT approach adds random RGB colors to mimic the full-color style in the unseen source (SVHN) without changing the content. 
}
\label{fig:generation_results_digits}
\end{figure}

\begin{figure}
\centering
    \begin{subfigure}[b]{\linewidth}
    \centering
        \includegraphics[width=\textwidth]{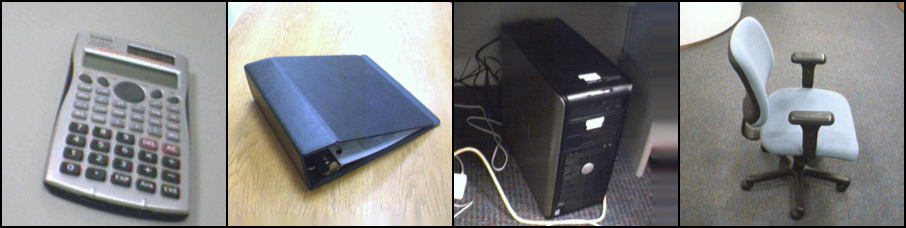}
        \caption{Target images (Webcam)}
    \end{subfigure}
    \begin{subfigure}[b]{\linewidth}
    \centering
        \includegraphics[width=\textwidth]{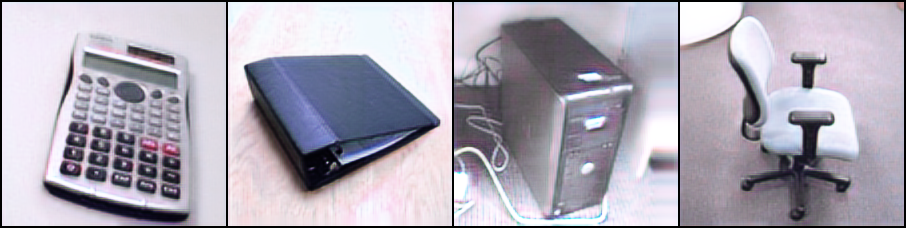}
        \caption{Generated source-style images}
    \end{subfigure}
    \begin{subfigure}[b]{\linewidth}
    \centering
        \includegraphics[width=\textwidth]{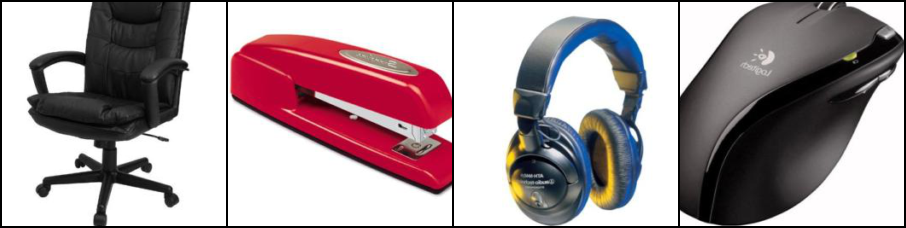}
        \caption{Unseen source images (Amazon)}
    \end{subfigure}
\caption{
Results from the SFIT method on the Office-31 dataset Amazon$\rightarrow$Webcam. Our translation method whitens backgrounds while increasing contrast ratios of the object (Webcam) for more appealing appearances as in the online shopping images (Amazon).
}
\label{fig:generation_results_office}
\end{figure}

\begin{figure*}[t]
\centering
    \begin{subfigure}[b]{\linewidth}
    \centering
        \includegraphics[width=\textwidth]{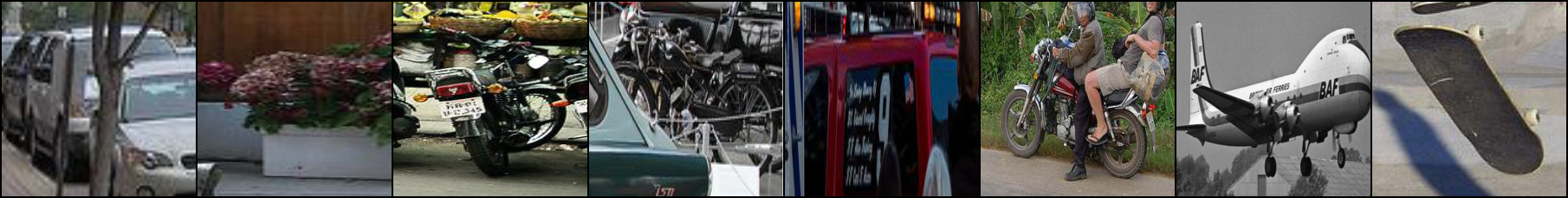}
        \caption{Target images (real-world)}
    \end{subfigure}
    \begin{subfigure}[b]{\linewidth}
    \centering
        \includegraphics[width=\textwidth]{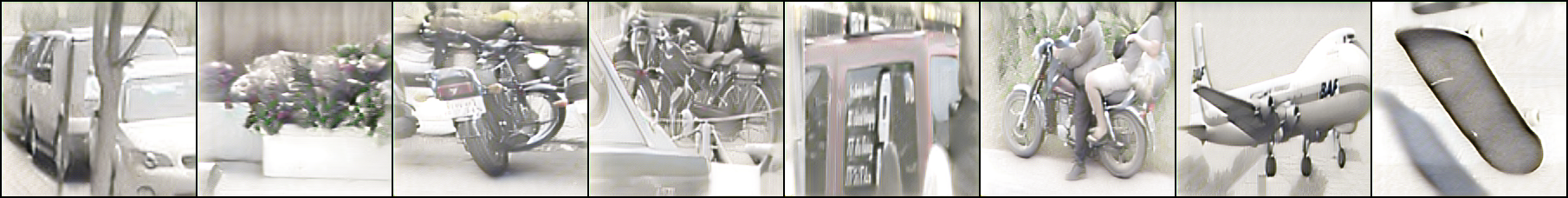}
        \caption{Generated source-style images}
    \end{subfigure}
    \begin{subfigure}[b]{\linewidth}
    \centering
        \includegraphics[width=\textwidth]{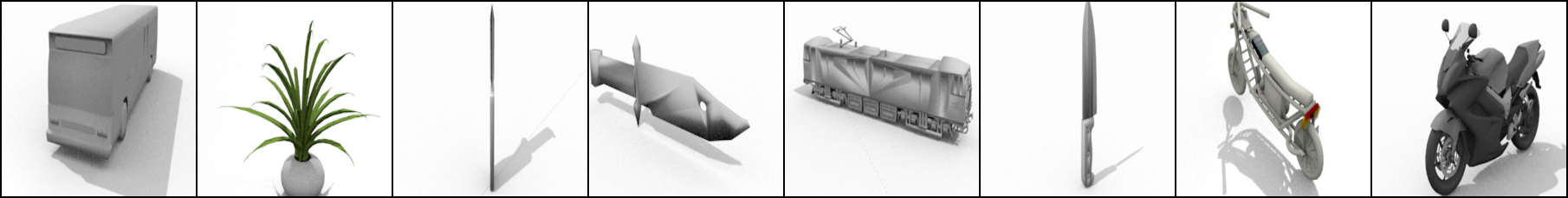}
        \caption{Unseen source images (synthetic)}
    \end{subfigure}
\caption{
Results from the SFIT method on the VisDA dataset SYN$\rightarrow$REAL. Our translation method decreases the target (real-world) image saturation and whitens the background while keeping the semantics unchanged. 
}
\label{fig:generation_results_visda}
\end{figure*}

\textbf{Recognition accuracy on generated images.}
To examine whether the proposed SFIT method can depict the knowledge difference, 
in Table \ref{tab:digits}-\ref{tab:visda}, we report recognition results using the generated-image-source-model branch (referred as ``generated images''). 
On the digits dataset, 
in terms of performance gaps, the knowledge differences between source and target models are 26.5\% on SVHN$\rightarrow$MNIST, 7.6\% on USPS$\rightarrow$MNIST, and 25.2\% on MNIST$\rightarrow$USPS. Generated images from SFIT bridges these differences to 0.2\%, 0.7\%, and 0.3\%, respectively.  
On the Office-31 dataset, 
the performance gap between the two models is 8.0\% on average, and the generated images shrink this down to 1.7\%. 
Notably, the performance drops from the target-image-target-model branch to the generated-image-source-model branch are especially pronounced on D$\rightarrow$A and W$\rightarrow$A, two settings that transfer Amazon images with white or no background to real-world background in Webcam or DSLR. In fact, in experiments we find generating an overall consistent colored background is very demanding, and the system usually generates a colored background around the outline of the object.   
On the VisDA dataset, 
generated images bridge the performance gap from 33.9\% to 6.9\%, even under a more demanding setting and a larger domain gap going from real-world images to synthetic CAD model images. 
Overall, on all three datasets, generated images significantly mitigate the knowledge difference in terms of performance gaps, indicating that the proposed SFIT method can successfully distill the adapted knowledge from the target model to the generated images.

\textbf{Visualization of source-free image translation results.}
For digits datasets SVHN$\rightarrow$MNIST (Fig.~\ref{fig:generation_results_digits}), the generator learns to add RGB colors to the gray-scale MNIST (target) images, which mimics the full-color SVHN (source) images. 
For Office-31 dataset Amazon$\rightarrow$Webcam (Fig.~\ref{fig:generation_results_office}), the generated images whiten the background, while having a white or no background rather than real-world background is one of the main characteristics of the Amazon (source) domain when compared to Webcam (target). Moreover, Amazon online shopping images also have higher contrast ratios for more appealing appearances, and our translated images also capture these characteristics, \eg, keys in the calculator, case of the desktop computer. 
For VisDA dataset SYN$\rightarrow$REAL (Fig.~\ref{fig:intro} and  Fig.~\ref{fig:generation_results_visda}), the generator learns to decrease the overall saturation of the real-world (target) objects which makes them more similar to the synthetic (source) scenario, while at the same time whitens the background, \eg, horse, truck, and plane in Fig.~\ref{fig:intro}, car and skateboard in Fig.~\ref{fig:generation_results_visda}, and brings out the green color in the plants. 
Overall, image generation results exhibit minimal content changes from target images, while successfully capturing the \textit{unseen} source style. 

In terms of visual quality, 
it is noteworthy that generation results for digits datasets SVHN$\rightarrow$MNIST contain colors and patterns that are not from the source domain, whereas our results on the Office-31 dataset and VisDA dataset are more consistent with the unseen source. Due to the lack of source images, rather than traditional image translation approaches~\cite{gatys2016image,huang2017arbitrary,pmlr-v80-hoffman18a,taigman2016unsupervised}, SFIT only relies on source and target models, and portrays adapted knowledge according to the two models. Since a weaker LeNet classifier is used for the digits dataset, it is easier to generate images that satisfy the proposed loss terms without requiring the generated images to perfectly mimic the source style. On Office-31 and VisDA datasets, given stronger models like ResNet, it is harder to generate images that can satisfy the loss terms. Stricter restrictions and longer training time lead to generation results more coherent with \textit{unseen} source images that also have better visual quality. 

\begin{figure}[]
\centering
    \begin{subfigure}[b]{0.24\linewidth}
    \centering
        \includegraphics[width=\textwidth]{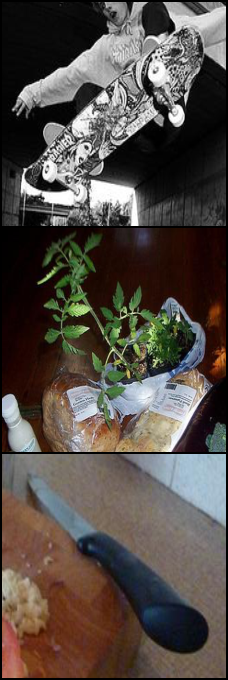}
        \caption{}
    \end{subfigure}
    \begin{subfigure}[b]{0.24\linewidth}
    \centering
        \includegraphics[width=\textwidth]{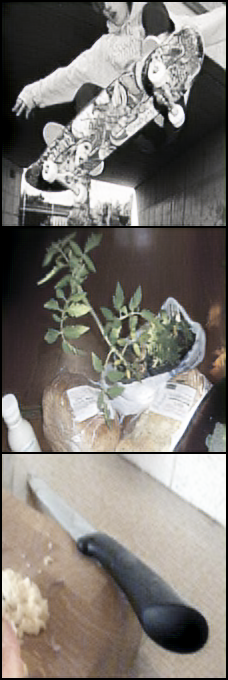}
        \caption{}
    \end{subfigure}
    \begin{subfigure}[b]{0.24\linewidth}
    \centering
        \includegraphics[width=\textwidth]{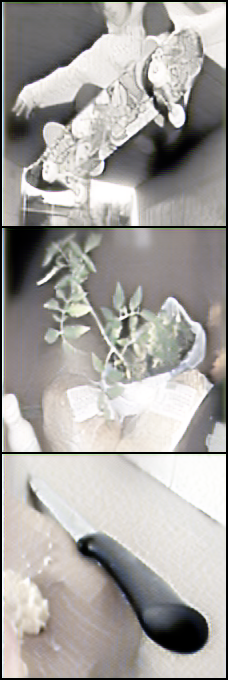}
        \caption{}
    \end{subfigure}
    \begin{subfigure}[b]{0.24\linewidth}
    \centering
        \includegraphics[width=\textwidth]{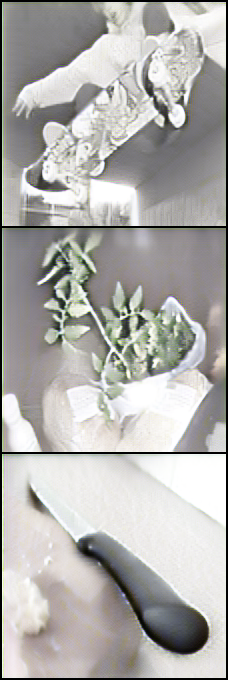}
        \caption{}
    \end{subfigure}
\caption{
SFIT results on VisDA dataset with different UDA methods. (a) Target images; (b) DAN~\cite{long2015learning}; (c) ADDA~\cite{tzeng2017adversarial}; (d) SHOT-IM~\cite{liang2020shot}. 
}
\label{fig:uda_methods}
\end{figure}

\textbf{Visualization for different UDA methods.}
In Fig.~\ref{fig:uda_methods}, we show SFIT visualization results using different UDA methods. 
Given source and target domain, a traditional image translation method generates a certain type of images regardless of the UDA methods, indicating its incapability of presenting the knowledge difference between models. In contrast, the proposed SFIT method generates different images for different UDA methods. Specifically, when comparing visualization results of the adapted knowledge in DAN~\cite{long2015learning}, ADDA~\cite{tzeng2017adversarial}, and SHOT-IM~\cite{liang2020shot}, we find stronger UDA methods can better transfer the target style to the unseen source style. As shown in Fig.~\ref{fig:uda_methods}, in terms of whitening the background for style transfer, SFIT results on ADDA are less coherent than SHOT-IM but better than DAN. This further verifies that our SFIT method indeed visualizes the knowledge difference between models, and stronger adaptation methods can better endure the style difference (leading to larger knowledge difference and thus stronger style transfer results).

\subsection{Application}
\label{sec:sec:applications}

The generated images from SFIT allows for further tuning of the target model in SFDA systems, where no source image is available. We include a diversity loss on all training samples to promote even class-wise distributions,
\begin{equation}
\label{eq:diversity}
    \mathcal{L}_\text{div} = -\mathcal{H}\left(\mathbb{E}_{\bm{x}\sim P_\text{target}\left(\bm{x}\right)}\left[p\left(f_\text{T}\left(\bm{x}\right)\right)\right]\right),
\end{equation}
where $\mathcal{H}\left(\cdot\right)$ denotes the information entropy function. We also incluse a pseudo-label fine-tuning loss, if pseudo label $\hat{y}_\text{S} = \argmax{p\left(f_\text{S}\left(\widetilde{\bm{x}}\right)\right)}$ from the generated-image-source-model branch equals to the pseudo label $\hat{y}_\text{T} = \argmax{p\left(f_\text{T}\left(\bm{x}\right)\right)}$ from the target-image-target-model branch. We then use this pseudo label $\hat{y}=\hat{y}_\text{S}=\hat{y}_\text{T}$ to fine-tune the target model,
 \begin{align}
\label{eq:ft}
\mathcal{L}_\text{pseudo}=\begin{cases}
               \mathcal{H}\left(p\left(f_\text{T}\left(\bm{x}\right)\right), \hat{y}\right), \;\;\;\; &\text{if}\;\; \hat{y}=\hat{y}_\text{S}=\hat{y}_\text{T}, \\
                0, \;\;\;\; &\text{else},
            \end{cases}
\end{align}
where $\mathcal{H}\left(\cdot,\cdot\right)$ denotes the cross entropy function. 
We combine these two loss terms in Eq.~\ref{eq:diversity} and Eq.~\ref{eq:ft} to give an overall fine-tuning loss $\mathcal{L}_\text{FT} =\mathcal{L}_\text{div} + \mathcal{L}_\text{pseudo}$. 

As an additional cue, supervision from generated-image-source-model further boosts target model SFDA performance. On Office-31, fine-tuning brings a performance improvement of 0.4\% according to Table~\ref{tab:office}. On VisDA, fine-tuning improves the target model accuracy by 0.7\% as shown in Table~\ref{tab:visda}. These improvements are statistically very significant (\ie, \textit{p}-value~$<$~0.001 over 5 runs), and introduce a real-world application for images generated by SFIT. 


\begin{figure}
\centering
    \begin{subfigure}[b]{0.24\linewidth}
    \centering
        \includegraphics[width=\textwidth]{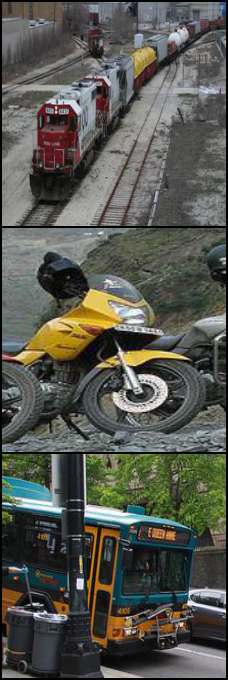}
        \caption{}
    \end{subfigure}
    \hfill
    \begin{subfigure}[b]{0.24\linewidth}
    \centering
        \includegraphics[width=\textwidth]{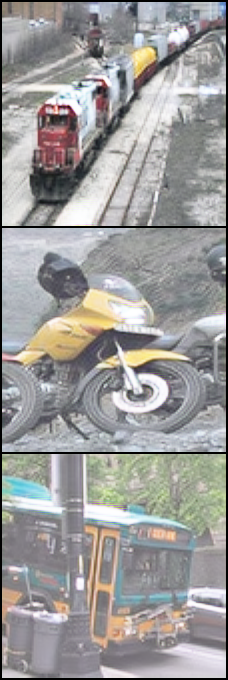}
        \caption{}
    \end{subfigure}
    \hfill
    \begin{subfigure}[b]{0.24\linewidth}
    \centering
        \includegraphics[width=\textwidth]{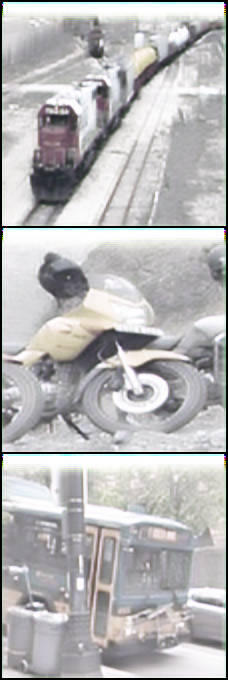}
        \caption{}
    \end{subfigure}
    \hfill
    \begin{subfigure}[b]{0.24\linewidth}
    \centering
        \includegraphics[width=\textwidth]{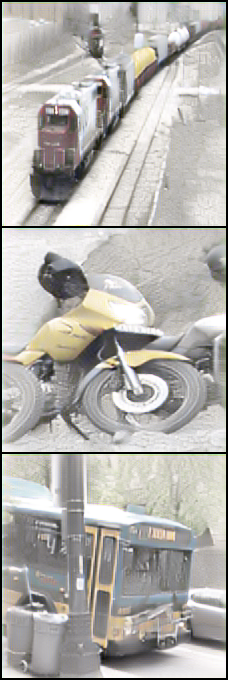}
        \caption{}
    \end{subfigure}
\caption{
Visualization results on VisDA dataset with different distribution alignment methods. (a) Target images; (b) BN stats alignment~\cite{hou2020source}; (c) traditional style loss~\cite{gatys2016image}; (d) relationship preserving loss. 
}
\label{fig:generation_variants}
\end{figure}

\subsection{Comparison and Variant Study}
\label{sec:sec:variants}

\textbf{Comparison with the BatchNorm statistics alignment method~\cite{hou2020source}.}
Hou~\etal propose to match the batch-wise feature map statistics so as to directly generate images that mimic the source style. Specifically, they explore the BatchNorm (BN) statistics stored in the BN layers in the source model for style indication, and match them against that of the generated images.
Using their approach, we can mildly change the image to the unseen source style (see Fig.~\ref{fig:generation_variants}) and slightly reduce the performance difference between the two branches (see Table~\ref{tab:generation_variants}). With that said, their lack of output alignments between the \textit{two branches} (only supervisions from the \textit{source branch}) results in much lower quantitative performance and under-performing style transfer quality when compared to the proposed method.

\textbf{Effect of the knowledge distillation loss.}
The knowledge distillation loss transfers the adapted knowledge to the generated images, and the removal of it results in a 1.1\% performance drop. 

\textbf{Effect of the relationship preserving loss.} As shown in Fig.~\ref{fig:generation_variants}, the traditional style loss can successfully transfer the target image to the source style on its own. However, using it causes a 4.8\% performance drop compared to the ``w/o $\mathcal{L}_\text{RP}$'' variant (see Table~\ref{tab:generation_variants}), suggesting it being unsuitable for SFIT. On the other hand, the batch-wise or pixel-wise relationship preserving variants \cite{tung2019similarity,li2020semantic} are found not useful, as they fail to improve over the ``w/o $\mathcal{L}_\text{RP}$'' variant. 

In contrast, the proposed channel-wise relationship preserving loss $\mathcal{L}_\text{RP}$ can effectively improve the recognition accuracy on the generated images, as the inclusion of it leads to a 2.6\% performance increase. 
Moreover, as shown in Fig.~\ref{fig:generation_variants}, similar to the traditional style loss, using only the relationship preserving loss can also effectively transfer the target image to the \textit{unseen} source style. 
Besides, focusing on the \textit{relative} channel-wise relationship instead of the \textit{absolute} correlation values, the proposed relationship preserving loss can better maintain the foreground object (less blurry and more prominent) while transferring the overall image style, leading to higher recognition accuracy. 


\begin{table}[t]
\centering
\small
\begin{tabular}{ccccc}
\toprule
Variant & $\mathcal{L}_\text{KD}$  &   $\mathcal{L}_\text{RP}$ & accuracy (\%) \\ \hline
Target image & -  & - & 46.8 \\
Initialized $g\left(\cdot\right)$ &   &   & 44.9 \\
BN stats alignment~\cite{hou2020source} &   &  & 51.7 \\
w/o $\mathcal{L}_\text{KD}$ & &  \cmark   & 72.7 \\
w/o $\mathcal{L}_\text{RP}$ & \cmark &  & 71.2 \\
$\mathcal{L}_\text{RP}\rightarrow\mathcal{L}_\text{style}$ & \cmark  & $\mathcal{L}_\text{style}$~\cite{gatys2016image} & 66.4 \\
$\mathcal{L}_\text{RP}\rightarrow\mathcal{L}_\text{batch}$ & \cmark  & $\mathcal{L}_\text{batch}$~\cite{tung2019similarity} & 71.2 \\
$\mathcal{L}_\text{RP}\rightarrow\mathcal{L}_\text{pixel}$ & \cmark  & $\mathcal{L}_\text{pixel}$~\cite{li2020semantic} & 70.9 \\
SFIT & \cmark  & \cmark & 73.8 \\
\bottomrule
\end{tabular}
\caption{Variant study on VisDA dataset. ``Initialized $g\left(\cdot\right)$'' refers to our transparent filter initialization in Section~\ref{sec:sec:implementations}.}
\label{tab:generation_variants}
\end{table}

\section{Conclusion}

In this paper, we study the scientific problem of visualizing the adapted knowledge in UDA. Specifically, we propose a source-free image translation (SFIT) approach, which generates source-style images from original target images under the guidance of source and target models. Translated images on the source model achieve similar results as target images on the target model, indicating a successful depiction of the adapted knowledge. Such images also exhibit the source style, and the extent of style transfer follows the performance of UDA methods, which further verifies that stronger UDA methods can better address the distribution difference between domains. We show that the generated images can be applied to fine-tune the target model, and might help other tasks like incremental learning. 


\section*{Acknowledgement}
This work was supported by the ARC Discovery Early Career Researcher Award (DE200101283) and the ARC Discovery Project (DP210102801).

{\small
\bibliographystyle{ieee_fullname}
\bibliography{egbib}
}

\end{document}